\documentclass[runningheads]{llncs}
\usepackage{graphicx}
\usepackage{xcolor}
\usepackage{hyperref}
\graphicspath{{./Images/}} 
\begin{document}
\title{Enhancing User's Income Estimation with Super-App Alternative Data}
%
%

\author{Gabriel Suárez\inst{1}\and
Juan Raful\inst{1} \and
Maria A. Luque\inst{1}\and
Carlos F. Valencia\inst{2}\and
Alejandro Correa-Bahnsen \inst{1}
}
\authorrunning{F. Author et al.}
%
\institute{Rappi, Cl. 93 \#19-58, Bogotá, Colombia.
\and Industrial Engineering, University of Los Andes, Cra 1 Este 19‑40, Bogota, Colombia
}

\maketitle              
\begin{abstract}
This paper presents the advantages of alternative data from Super-Apps to enhance user's income estimation models. It compares the performance of these alternative data sources with the performance of industry-accepted bureau income estimators that takes into account only financial system information; successfully showing that the alternative data manage to capture information that bureau income estimators do not. By implementing the TreeSHAP method for Stochastic Gradient Boosting Interpretation, this paper highlights which of the customer’s behavioral and transactional patterns within a Super-App have a stronger predictive power when estimating user’s income. Ultimately, this paper shows the incentive for financial institutions to seek to incorporate alternative data into constructing their risk profiles.

\end{abstract}
\section{Introduction}
Technology-based companies are disrupting traditional business lines while they show an unprecedented ability to capture, manage, and analyze large volumes of customer data, previously unavailable to the organizations that traditionally operated in such lines, and therefore becoming
increasingly competitive within them \cite{Einav2014,Shaw2014,McAfee2012}. Given this, the business model of Super-Apps has emerged. Those are, mobile applications that seek to satisfy a various number of customer’s daily needs, all within the same marketplace-based application. Whereas traditionally, users could only found those different services and solutions in a mobile application specifically designed for providing each one of them, like Uber for transportation and Uber Eats for delivery services.

One of the traditional sectors Super-Apps is venturing into is the financial sector, and so, it is logical then to raise the question on whether this Super-App data adds value within this field. This publication will explore how this alternative data might help financial institutions address one of the fundamental questions in credit risk management; that is, how much it should be lent to a customer, by building the most accurate user income estimate. This paper will then try to determine the following research questions:

\begin{enumerate}
\item Is there a significant improvement in income estimation statistical performance when using alternative data sources, retrieved from a Super-App, compared to industry-accepted Income Estimates?
\item What behaviors these Super-App features reveal, and how do they differ from traditional financial information resources?
\item Which of these behaviors appear to offer more predictive power?
\end{enumerate}

\section{Credit Risk and Income Estimation}

Financial Institutions have to effectively manage credit risk to lend money to their customers, being this the potential that their counter party will fail to meet its obligations under agreed terms \cite{Basel2001}. Hence, the challenge for these companies relies on the fact that each customer has a different probability of failing his particular obligations (PD) (as this is the reflection of each customer individual psychological characteristics \cite{tokunaga1993} together with its financial knowledge, demographic characteristics and situational factors \cite{perry2008})
, as well as an individualized credit loss amount if the debtor of the loan defaults or so called loss given default (LGD) \cite{tokunaga1993,perry2008,Basel2001,schuermann2004}.
Therefore, financial institutions have to answer two fundamental questions for their accurate credit risk management, being one to which customers should these institutions lend money to and the second one, how much money should they lend to them. To address the second question, it is necessary for financial institutions to estimate each of its customers' payment capability. Properly calculating the amount of money they should lend to its customers allows financial institutions to accurately estimate their exposure at default (EAD)  \cite{Basel2001}, which ultimately leads to organizations to operate within their desired expected loss (EL) in correspondence with their risk appetite. $ EL = PD * LGD * EAD. $

Traditionally, financial institutions take user's income as a starting point to answer this second question. Nevertheless, the willingness of consumers to provide this information is relatively low, as  user's income, being part of the user's financial information, is the most sensitive data to provide from  their personal identifiable information \cite{phelps2000}. This, leads financial institutions to try estimate user's income from their available financial information, and in some cases having to require it from third party institutions that gather this type of data. This estimate is, for some organizations, the best proxy for knowing this relevant user feature. However,  having an income estimator that relies absolutely on financial information marginalizes the opportunity to have a complete  profile of  those users who do not have information within the financial system.

\section{Users Interactions with Super-Apps}

Alternative Data sourced from a Super-App can be retrieved from the various ways users of the Super-App interact and navigate through it. Moreover, each transaction carried out within them different solutions (Delivery of restaurants/groceries/goods, transportation, travel, e-commerce and many more) generates different types of data. The features that can be collected then variate in their values in unique ways for each user as they reflect their individual behavioral and consumption patterns, and can be grouped into four categories:

\textbf{Personal Information:} The identifiable demographic attributes of the user such as age, place of residence, the brand of cell phone as well as personally identifiable information such as wealth estimators or address from where the orders where placed. 

\textbf{Consumption Patterns:} User's consumption patterns are retrieved from the delivery vertical within the Super-App, which are all the services and solutions related to the purchase and delivery of groceries, food, clothing, technology, pharmaceutical products, and others.

\textbf{Payment Information:} This is the information that can be retrieved from all of the transactions the user makes in the Super-App. The features that can be extracted from this type of interaction allow identifying user's favorite payment method, tendencies of installments when paying regular orders versus more expensive orders, number of times a credit card is declined and even the number of credit cards the user has available to pay within the app and the level of them.

\textbf{Financial services:} This last set of features comes from the fintech functionality within the Super-App. These features collect users' behavior towards the financial services or products delivered via technology ranging from e-wallets and digital cards to loan services, on- and offline payments, and money transfers \cite{Roa2020}.

\section{Experimental Setup}

\subsection{Data}

Our data set consists of the transactional historical information from 43.270 users within a Super-App. This transactional information is collected directly from the orders and interactions with more than 15.000 restaurants and 2.000 grocery stores. From these orders, it was possible to retrieve several other features such as the latitude and longitude from which each order was placed, time features such as the day of the week and the hour where the order was made, the used payment method and - when applicable - its credit card information, the device and the operating system used to interact with the super-app and many more. All this data allowed us to understand each customer's consumption patterns and build the variables that might be indicators of each user's income. Moreover, we had access to the real-validated income of the users, who built the training and testing sets for the model, together with their Bureau Income Estimates.

\subsection{Setup}

Observing the existing research in terms of credit risk and income estimation, and in order to make the most of the available data and the computational power of artificial intelligence, it was decided to implement an XGBoost model for the scope of this research as it has proven to outperform other algorithms such as neural networks, support vector machines, bagging-NN and many others with regard to structured data \cite{Xia2017,Salvaire2019}. And furthermore has showed predictive power for credit risk assessment models using alternative data. \cite{Roa2020}

Considering this research is being carried out within a business environment, evaluating statistical model performance has to come in handy in terms of interpretability to discuss its possible business implications effectively. Hence, we will evaluate the proposed model according to the Mean Average Percentual Error (MAPE). This metric is a measure of prediction accuracy and a widely accepted indicator within the organization environment when considering the utilization of a prediction method \cite{khair2017}.  This metric is also a variation of the Mean Absolute Error used to evaluate previous Income Estimate models' performance in literature \cite{Pietro2015}. We implemented cross-validation with five splits for evaluating the model, keeping a data proportion of 80\% -20\% for training and testing in each iteration, respectively. Furthermore,we conducted statistical tests such as Levene for homoscedasticity of variables and Mann-Whitney to evaluate of significance in the MAPE.

The Mean Average Percentual Error (MAPE) used for the performance evaluation, is no more that the average of each observation of the percentual error of it, MAPE can be expressed as :

    $$ \mbox{MAPE} = \frac{1}{n}\sum\limits_{i=1}^n  \frac{A_i - F_i}{A_i} $$

Where $ A_i$ indicates the actual value of the user's income for the observation $ i $ while $F_i $ indicates the forecasted value of the user's income for the observation $ i $.

\section{Results}

\subsection{Statistical model performance}

\begin{figure}[h!]
  \centering
  \begin{minipage}[b]{0.9\textwidth}
    \includegraphics[width=\textwidth]{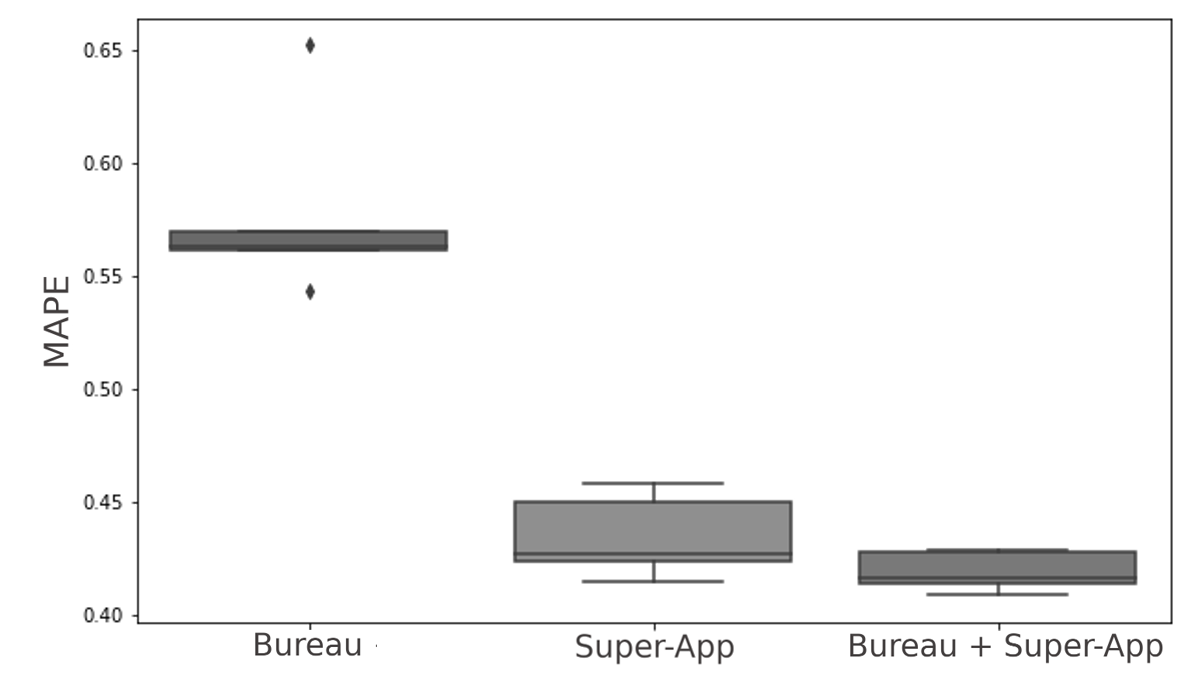}
    \caption{MAPE  performance by model}
    \label{mape}
  \end{minipage}
  \end{figure}

The results obtained in this study show a significant improvement in performance in terms of the MAPE for the model that incorporates Super-App alternative data compared with the Bureau Income Estimates by themselves as can be seen in Fig.~\ref{mape} and is confirmed by the result for the p-value of the Mann Whitney non-parametric test for both Bureau vs. Super-App and Bureau + Super-App vs Bureau comparison in Table 1.  Moreover, an equivalent performance of the combined model versus the Super-App Only is obtained, according to the p-value of the Mann-Whitney non-parametric test results, presented in Table 1. The p-values of the Levene test conducted to verify homoscedasticity and hence the applicability of the Mann Whitney test, are also presented in Table \ref{Mean_test_EMP}.

Therefore, our results suggest that financial institutions can build better and more accurate estimates solely with this alternative data, without requiring a particular bureau income estimate. This, as the model that is built with both alternative data and the Bureau Income Estimate (Bureau + Super-App) does not outperforms the alternative data-only model (Super-App) Fig. \ref{mape} supported by equivalent result of the Mann Whitney  non-parametric test Table \ref{Mean_test_EMP}. Which ultimately, indicates that the bureau income estimate subject of this study does not manages to capture any significant information that the Super-App alternative data is not already capturing in terms of the user's creditworthiness. It is worth mention that building a particular model only with the retrievable features that the interaction of a user with a Super-App generates theoretically does not line up with the stated objective of credit risk management, that is to incorporate all the available information of a user to build the most detailed and truthful profile possible. However, the results indicate that the bureau income estimate does not add value when building this profile to assets income estimate, and the Super-App features are showed to be sufficient to complete said profile without having to resort to the financial history of the users. Ultimately, pointing out that financial institutions, that manage to incorporate this alternative data when building their credit risk profiles, can reduce the risk in their decision-making process of addressing how much money they should lend to their customers, operating under a more accurately estimated expected loss (EL).


  \begin{table}[h!]
\centering
\caption{Mann Whitney and Levene test P-Value for MAPE performance}
\begin{tabular}{lcccc}

Model Comparison    & Levene  & Mann Whitney   \\

Bureau, Super-App      & 0.96364    & 0.00609  \\
Bureau + Super-App, Bureau & 0.40492    & 0.00609 \\
Bureau + Super- App, Super-App & 0.29555    & 0.14813  \\

\end{tabular}
\label{Mean_test_EMP}
\end{table}

\subsection{Feature Importance}
In order to be able to determine which of the behavioral patterns identified have more predictive power when estimating users income, as previously mention, we implemented SHapley Additive exPlanations \cite{shap2017}  in order to give some degree of explainability to the results of the executed XGBoost model. 

\begin{figure}[h!]
  \centering
  \begin{minipage}[b]{1\textwidth}
    \includegraphics[width=\textwidth]{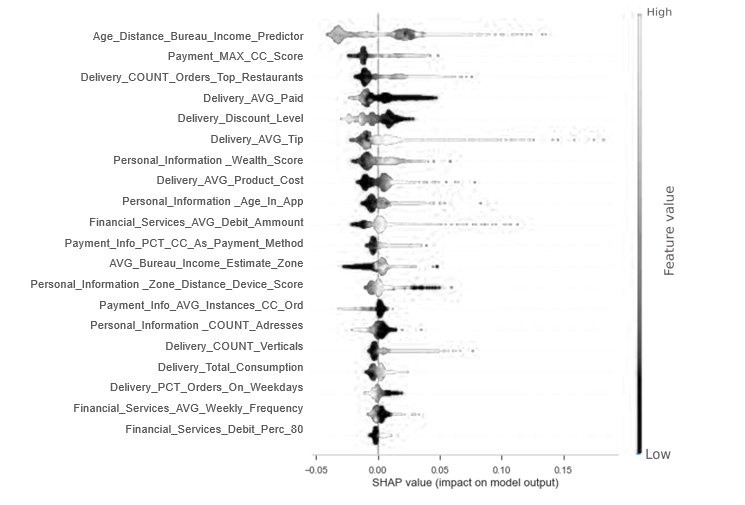}
    \caption{Local interpretability with SHAP of Bureau + Super-App model}
    \label{shap}
  \end{minipage}
\end{figure}

Figure~\ref{shap} is the visualization for the Bureau Income Estimates combined with Super-App alternative data model feature importance, where features appear from top to bottom in order of importance; Figure~\ref{shap} shows that the distance of the industry-accepted Bureau Income Predictor of the user and the average of the same estimate for the users within the same age range presents the highest predictive power from the considered features. In contrast, as for alternative data features, the number of times the user has placed orders at one of the most expensive restaurants within the Super-App (Where the average plate has a value higher than the average plate of 70\% of the other restaurants) has the most robust predictive value. Furthermore, Figure~\ref{shap} shows each feature's impact on the combined features model. Notably, users with a taste for expensive (top) restaurants (Delivery\_COUNT\_Orders\_Top\_Restaurants), a higher delivery consumption (Delivery\_Total\_Consumption), a higher amount of money debited through financial services (Financial\_Services\_AVG\_Debit\_Amount and Financial\_Services\_Debit\_Perc\_80), users with high-level or premium credit cards registered in the app (Payment\_MAX\_CC\_Score), users who differ to a large number of instances orders paid by credit card  (Payment\_Info\_AVG\_Instances\_CC\_Ord) and users with a preference for the Credit Card as payment method (Payment\_Info\_PCT\_CC\_As\_Payment\_Method) are estimated to have a higher income than their counterparts. Moreover, users with a high engagement towards the Super-App different verticals were also predicted with higher incomes than users with a lower engagement (Delivery\_Count\_Vertocales). It can also be highlighted that users with a taste for discounts are predicted to have lower incomes than users who do not use discounts in their delivery orders. As can be seen with the performance of variables like Delivery Discount Level' and 'Delivery AVG Paid,' both proxies of the user's behavior towards discounts, where a high values of the variable indicates high adoption towards discounts and, as can be seen in Figure ~\ref{shap}, are associated with low income users. The tipping behavior results show that users who on average tip less in their delivery orders are estimated to have lower incomes than users that leave higher tips. Hence our results indicate that the propensity of tipping high, the taste for expensive restaurants and products, credit card preference, short term period installments for credit card payments and overall a high monetary consumption and high amount of money debited through financial services are all strong positive income estimators. On the other hand, our results show that the preference to place orders with discounts is a robust negative income estimator.

\section{Conclusions}

This paper determined that the proposed Super-App alternative data have predictive power towards income estimation. Moreover, there is a significant improvement in statistical performance when taking these alternative data sources into account compared to the performance of industry-accepted Bureau Income Estimates by themselves. It was possible to identify consumption patterns, personal information, payment behaviors\_preferences, engagement towards financial products and engagement towards the Super-App verticals from extracting data from the users' interactions with the Super-App. These behaviours presented by alternative data showed predictive power when estimating user's income, being all this information non-available to traditional bureau estimates, that incorporate only past financial information of the user.

Moreover, we presented how financial institutions that possess this kind of information should include it into building their users' credit risk profiles.; as this would implicate operating within a more accurate expected loss. Furthermore, this paper layout how financial institutions that manage to incorporate Super-App sourced alternative information into their credit risk profiling will be able to assess the income of customers who do not have previous financial information which should represent a benefit towards the bankarization of this population, which is outside the reach of traditional bureau income estimates.

This study's results present the incentive for financial institutions to seek to incorporate this type of information into constructing their risk profiles. Doing so will allow them to keep up with these technology-based financial institutions that, in the meantime, will have a competitive advantage if they manage to align their strategical operation with the insights of the data they can collect. Additionally, our findings should motivate further research in this field to determine what other sources of alternative information can provide value towards building complete and accurate profiles for financial institutions' users.

\end{document}